\documentclass{article} 
\usepackage[preprint]{colm2026_conference}

\usepackage{microtype}
\usepackage[T1]{fontenc}
\usepackage{hyperref}
\usepackage{url}
\usepackage{booktabs}
\usepackage{amsmath}
\usepackage{amssymb}
\usepackage{multirow}
\usepackage{xcolor}
\usepackage{colortbl}
\usepackage{graphicx}
\usepackage{listings}
\usepackage{algorithm}
\usepackage{algpseudocode}
\lstdefinestyle{promptstyle}{
    basicstyle=\scriptsize\ttfamily,
    breaklines=true,
    breakatwhitespace=false,
    columns=fullflexible,
    keepspaces=true,
    frame=single,
    framerule=0.4pt,
    rulecolor=\color{black!40},
    backgroundcolor=\color{black!4},
    xleftmargin=4pt,
    xrightmargin=4pt,
    aboveskip=8pt,
    belowskip=8pt,
    language={},
}
\usepackage{lineno}
\usepackage{tikz}
\usepackage{pgfplots}
\pgfplotsset{compat=1.18}
\usetikzlibrary{
    arrows.meta,
    positioning,
    shapes.misc,
    shapes.geometric,
    fit,
    backgrounds,
    calc,
    decorations.pathreplacing,
    patterns,
    shadows.blur,
    pgfplots.groupplots
}

\definecolor{darkblue}{rgb}{0, 0, 0.5}
\definecolor{lightblue}{rgb}{0.88, 0.93, 1.0}
\definecolor{umassmaroonlight}{RGB}{255, 232, 232}
\definecolor{umassmaroon}{RGB}{136, 28, 28}
\definecolor{maroonlight}{RGB}{245, 215, 215}
\definecolor{maroonbg}{RGB}{255, 242, 242}
\definecolor{accentblue}{RGB}{41, 98, 155}
\definecolor{bluelight}{RGB}{210, 228, 250}
\definecolor{bluebg}{RGB}{235, 245, 255}
\definecolor{fixedgray}{RGB}{225, 225, 225}
\definecolor{textgray}{RGB}{110, 110, 110}
\definecolor{darktext}{RGB}{40, 40, 40}
\definecolor{greenbg}{RGB}{220, 245, 220}
\definecolor{greentext}{RGB}{30, 120, 50}
\definecolor{orangebg}{RGB}{255, 240, 215}
\definecolor{orangetext}{RGB}{180, 100, 20}
\definecolor{amberdark}{RGB}{175, 125, 25}
\definecolor{amberlight}{RGB}{255, 236, 179}
\definecolor{oraclegreen}{RGB}{46, 139, 87}
\definecolor{rankerpurple}{RGB}{108, 52, 168}
\definecolor{rankerpurplelight}{RGB}{232, 218, 250}
\definecolor{deltared}{RGB}{200, 40, 40}
\definecolor{deltagreen}{RGB}{30, 120, 50}
\definecolor{baselinegray}{RGB}{232, 232, 242}

\newcommand{\flameicon}[1][0.9em]{\raisebox{-0.15em}{\includegraphics[height=#1]{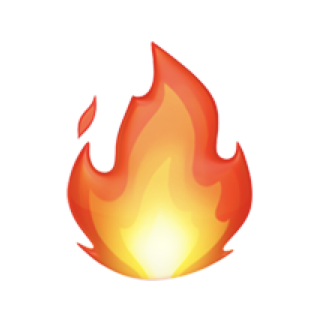}}}
\newcommand{\iceicon}[1][0.9em]{\raisebox{-0.15em}{\includegraphics[height=#1]{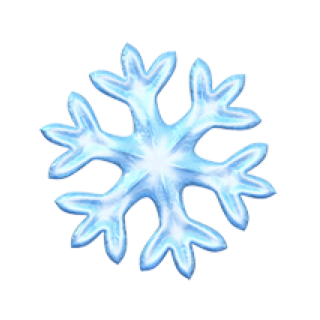}}}
\hypersetup{colorlinks=true, citecolor=darkblue, linkcolor=darkblue, urlcolor=darkblue}

\newcommand{\methodname}{\textsc{CoSearch}}

\title{CoSearch: Joint Training of Reasoning and Document Ranking via Reinforcement Learning for Agentic Search}

\author{Hansi Zeng$^1$, Liam Collins$^2$, Bhuvesh Kumar$^2$, Neil Shah$^2$, Hamed Zamani$^1$ \\
$^1$ Center for Intelligent Information Retrieval, University of Massachusetts Amherst\\
$^2$ Snap Inc.
}

\begin{document}

\ifcolmsubmission
\linenumbers
\fi

\maketitle

\begin{abstract}
Agentic search -- the task of training agents that iteratively reason, issue queries, and synthesize retrieved information to answer complex questions --  has achieved remarkable progress through reinforcement learning (RL).  However, existing approaches such as Search-R1 treat the retrieval system as a fixed tool, optimizing only the reasoning agent while the retrieval component remains unchanged.
A preliminary experiment reveals that the gap between an oracle and a fixed retrieval system reaches up to \textbf{+26.8\%} relative F1 improvement across seven QA benchmarks, suggesting that the retrieval system is a key bottleneck in scaling agentic search performance.
Motivated by this finding, we propose \methodname{}, a framework that jointly trains a multi-step reasoning agent and a generative document ranking model via Group Relative Policy Optimization (GRPO). To enable effective GRPO training for the ranker---whose inputs vary across reasoning trajectories---we introduce a semantic grouping strategy that clusters sub-queries by token-level similarity, forming valid optimization groups without additional rollouts. We further design a composite reward combining ranking quality signals with trajectory-level outcome feedback, providing the ranker with both immediate and long-term learning signals. Experiments on seven single-hop and multi-hop QA benchmarks demonstrate consistent improvements over strong baselines, with ablation studies validating each design choice. Our results show that joint training of the reasoning agent and retrieval system is both feasible and strongly performant, pointing to a key ingredient for future search agents.
Code is available at \url{https://github.com/snap-research/CoSearch}
\end{abstract}

\section{Introduction}
\label{sec:introduction}

Large language models (LLMs) have demonstrated strong reasoning capabilities, yet they face fundamental challenges when questions require knowledge beyond their parametric training data. Retrieval-augmented generation (RAG)~\citep{rag,REML} addresses this limitation by allowing models to access external knowledge through retrieval. While early approaches relied on single-turn retrieval followed by answer generation, recent work has shown that multi-step agentic search---where models iteratively reason, generate search queries, and incorporate retrieved information---yields substantially stronger performance on knowledge-intensive tasks~\citep{ir-cot,baleen,shi-etal-2024-generate}. In these agentic search systems, an agent interacts with a retrieval system iteratively: the agent reasons about what information is needed, issues a search query, observes the retrieved documents, and repeats this process until sufficient evidence is gathered to produce a final answer. Training such agents through reinforcement learning (RL), where the reward is based on final answer correctness, has proven effective at teaching them \emph{when} and \emph{what} to search~\citep{search-r1,webthinker,zeng2026synplanresearchr1encouragingtoolexploration}. Despite this progress, a common assumption across these systems is that the retrieval model or search engine remains unchanged. This means that the agent treats the retrieval model as a tool and learns to use it and the tool does not adapt.

\begin{table}[t]
\centering
\small
\label{tab:oracle}
\setlength{\tabcolsep}{7pt}
\renewcommand{\arraystretch}{1.2}
\begin{tabular}{l ccc ccc}
\toprule
& \multicolumn{3}{c}{\textbf{7B Main Agent}} & \multicolumn{3}{c}{\textbf{3B Main Agent}} \\
\cmidrule(lr){2-4} \cmidrule(lr){5-7}
& Single & Multi & \textbf{Avg} & Single & Multi & \textbf{Avg} \\
\midrule
Standard & 0.600 & 0.506 & 0.546 & 0.531 & 0.371 & 0.440 \\
Oracle Retrieval & 0.709 & 0.564 & 0.626 & 0.636 & 0.499 & 0.558 \\
\midrule
\rowcolor{umassmaroonlight}
\textbf{Rel.\ Gain} & \textbf{+18.2\%} & \textbf{+11.5\%} & \textbf{+14.7\%} & \textbf{+19.8\%} & \textbf{+34.5\%} & \textbf{+26.8\%} \\
\bottomrule
\end{tabular}
\caption{Oracle retrieval gap (F1 score). Performance averaged over single-hop (PopQA, NQ, TQA) and multi-hop (HotpotQA, 2Wiki, Musique, Bamboogle) benchmarks.}
\end{table}

We argue that this assumption introduces a significant bottleneck.
Through analysis of RL-trained search agents, we observe that most errors originate not from the reasoning process but from the retrieval component: the agent issues queries that look reasonable, yet the retrieval system fails to return relevant documents, forcing the agent to either retry with reformulated queries or draw incorrect conclusions from noisy information. To quantify this gap, we conduct an oracle retrieval experiment: at each retrieval step, we promote all documents matching the gold (ground-truth) answer to the top positions; for sub-queries where no retrieved document matches the gold answer, the ranking is left unchanged (see Appendix~\ref{app:oracle} for details). 
As shown in Table~\ref{tab:oracle}, this simple intervention consistently improves average F1 by \textbf{+14.7\%} for a 7B-parameter agent and by \textbf{+26.8\%} for a 3B agent across seven QA benchmarks, demonstrating that the retrieval bottleneck holds broadly regardless of the agent's reasoning capacity. 
These findings motivate the following question: \emph{can we jointly optimize the retrieval system alongside the multi-turn reasoning agent, so that they learn to complement each other for providing accurate responses?}
In this work, we propose \methodname{}, a reinforcement learning (RL) framework that jointly trains a main reasoning agent and a generative ranker for agentic search tasks. The main agent operates in a standard ReAct-style loop~\citep{yao2023reactsynergizingreasoningacting}, generating thoughts and search queries. The generative ranker receives the main agent's sub-queries along with candidate documents from a fixed first-stage dense retriever, and selects and reranks a subset for the main agent to observe; together, the retriever and ranker form the retrieval system. The main agent and the ranker are trained simultaneously using Group Relative Policy Optimization (GRPO)~\citep{deepseek2025deepseekr1}, with rewards derived from the correctness of the final answer.

Applying GRPO to the ranker introduces two key technical challenges. First, GRPO computes advantages over groups of responses generated from the \emph{same input prompt}. While this is naturally satisfied for the main agent---where multiple trajectories are sampled from the same user query---the ranker receives different sub-queries across trajectories, preventing direct grouping. We address this through a \emph{semantic grouping} strategy that clusters sub-queries by token-level F1 similarity, forming valid GRPO groups without additional rollouts. Second, the ranker's effect on the final answer is indirect: good retrieval may still yield an incorrect answer if reasoning fails, and vice versa. We design a \emph{composite reward} combining a relevance reward based on a ranking quality metric (Hit@$k$) with a trajectory-level reward reflecting the final answer correctness, providing both immediate feedback on retrieval precision and long-term signal on downstream utility. Both reward components are computed using rule-based metrics, requiring no expensive LLM annotations.

Following prior work \citep{search-r1}, we evaluate \methodname{} on seven QA benchmarks spanning single-hop (PopQA, NQ, TriviaQA) and multi-hop (HotpotQA, 2WikiMultiHopQA, Musique, Bamboogle) settings. Using Qwen2.5-7B-Instruct as the backbone for both the main and ranking agents, our method consistently outperforms strong baselines including Search-R1 \citep{search-r1} and fixed ranker variants. We further demonstrate effectiveness with a 3B main agent paired with a 7B ranker. Our main contributions are as follows:
\begin{itemize}
    \item We identify retrieval quality as a critical bottleneck in RL-trained agentic search systems: an oracle retrieval experiment reveals relative F1 gains of +14.7\% for a 7B reasoning agent and +26.8\% for a 3B agent, demonstrating a substantial and broadly held performance ceiling imposed by the fixed retrieval system.


    \item We propose \methodname{}, a framework that jointly trains a reasoning agent and a generative document ranker via RL. To enable this, we introduce a semantic grouping strategy for applying GRPO to the ranker despite varying inputs across trajectories, and a composite reward that combines ranking quality with trajectory-level answer correctness.

    \item Experiments on seven QA benchmarks show consistent improvements over strong baselines, achieving +6.6\% relative F1 over Search-R1 with a 7B agent and +10.8\% with a 3B agent. Our results show that jointly training the retrieval system and reasoning agent is not only feasible but strongly performant, and is likely a key ingredient for future search agents.
\end{itemize}

\section{Related Work}
\label{sec:related-work}

\paragraph{Agentic search and deep research.}
Retrieval-augmented generation~\citep{rag,realm,pmlr-retro,karpukhin-etal-2020-dense} enables LLMs to access external knowledge, but single-turn retrieval often fails on complex questions requiring multi-step evidence gathering. This has motivated agentic search systems that interleave reasoning with retrieval~\citep{react,ir-cot,selfrag}, with recent deep research agents~\citep{openai2025deepresearch,google2025geminideepresearch,Nakano2021WebGPTBQ} coupling large reasoning models with web search for complex queries. The advent of RLVR has further enabled training search agents directly from answer correctness~\citep{search-r1,search-o1,song2025r1searcher,webthinker,simpledeepsearcher,guan2025deepragthinkingretrievestep,zeng2026synplanresearchr1encouragingtoolexploration}. 

\paragraph{Training retrievers for LLMs.}
A separate line of work adapts retrieval for downstream LLM generation. Prior approaches train retrievers via language model supervision~\citep{realm,shi2024replug,sachan2021endtoend,Alireza2024uRAG}, while recent work shows that LLMs can serve as effective listwise rankers~\citep{Sun2023IsCG,Pradeep2023RankZephyrEA}. Most related, Agentic-R~\citep{Liu2026AgenticRLT} trains a dense retriever for agentic search via LLM-annotated labels and iterative contrastive learning. In contrast, \methodname{} uses a generative ranker trained end-to-end with RL in a simultaneous joint framework, requiring only lightweight rule-based rewards.

\paragraph{RL for tool-augmented LLMs.}
RLVR~\citep{openai2024learning,deepseek2025deepseekr1,lambert2024tulu3,kaufmann2025survey} has become a key paradigm for optimizing LLMs, with value-free methods such as GRPO~\citep{shao2024deepseekmath} and RLOO~\citep{ahmadian2024back} offering simpler alternatives to PPO~\citep{schulman2017proximalpolicyoptimizationalgorithms}. Applying RLVR to tool-augmented settings has attracted growing interest~\citep{toRL,retool,arpo,ragen,gigpo,xue2025simpletir,dong2025aepo}, but these works focus on training the \emph{reasoning agent} to use tools more effectively. In contrast, our work applies GRPO to train the \emph{tool itself}, specifically the rankers.
\section{Methodology}
\label{sec:methodology}

\begin{figure}[t]
\centering
\begin{tikzpicture}[
    >=Stealth,
    every node/.style={font=\sffamily},
    comp/.style={
        rounded corners=6pt, line width=0.9pt,
        minimum height=1.0cm, minimum width=1.9cm,
        align=center, font=\sffamily\small
    },
    dataflow/.style={->, line width=1pt},
    scale=1.05, transform shape
]

\fill[fixedgray!30, rounded corners=10pt] (-1.4, 0.15) rectangle (11.5, 3.0);
\node[font=\sffamily\small\bfseries, textgray] at (1.0, 2.7) {Existing: Single-Agent RL};

\node[comp, draw=umassmaroon, fill=maroonlight] (lmain) at (0.7, 1.5)
    {Main Agent \flameicon[0.7em]\\[-1pt]{\footnotesize (trainable)}};
\node[comp, draw=textgray, fill=fixedgray] (lret) at (3.9, 1.5)
    {Retriever \iceicon[0.7em]\\[-1pt]{\footnotesize (fixed)}};
\node[comp, draw=textgray, fill=fixedgray] (lrr) at (6.7, 1.5)
    {Ranker \iceicon[0.7em]\\[-1pt]{\footnotesize (fixed)}};

\draw[textgray!50, dashed, rounded corners=6pt, line width=0.7pt]
    (2.6, 0.55) rectangle (8.0, 2.5);
\node[font=\sffamily\tiny, textgray!70, anchor=south] at (5.3, 0.55) {Retrieval System};

\draw[dataflow, textgray] (lmain) -- node[above, font=\sffamily\tiny] {$q_t$} (lret);
\draw[dataflow, textgray] (lret) -- node[above, font=\sffamily\tiny] {$\mathcal{D}_N$} (lrr);
\draw[dataflow, textgray, rounded corners=4pt] (lrr.south) -- ++(0,-0.4) -| node[below, font=\sffamily\tiny, pos=0.18] {$o_t$ (top-$K$)} (lmain.south);

\node[font=\sffamily\scriptsize, textgray, align=center, anchor=west] (lrew) at (9.3, 1.5)
    {$r = \text{F1}(a,\, a^*)$};
\node[font=\sffamily\tiny, textgray, anchor=north] at (lrew.south) {GRPO $\to$ Main only};
\draw[dataflow, textgray, dashed] (lrr) -- (lrew);

\node[font=\sffamily\normalsize\bfseries, textgray] at (4.8, 0.0) {vs.};

\fill[bluebg, rounded corners=10pt] (-1.4, -3.1) rectangle (11.5, -0.2);
\node[font=\sffamily\small\bfseries, accentblue] at (0.6, -0.5) {\methodname{}};

\node[comp, draw=umassmaroon, fill=maroonlight] (rmain) at (0.7, -1.7)
    {Main Agent \flameicon[0.7em]\\[-1pt]{\footnotesize (trainable, $\pi_{\theta_{\text{main}}}$)}};
\node[comp, draw=textgray, fill=fixedgray] (rret) at (3.9, -1.7)
    {Retriever \iceicon[0.7em]\\[-1pt]{\footnotesize (fixed)}};
\node[comp, draw=rankerpurple, fill=rankerpurplelight] (rrr) at (6.7, -1.7)
    {Ranker \flameicon[0.7em]\\[-1pt]{\footnotesize (trainable, $\pi_{\theta_{\text{rr}}}$)}};

\draw[darktext!40, dashed, rounded corners=6pt, line width=0.7pt]
    (2.6, -2.7) rectangle (8.0, -0.7);
\node[font=\sffamily\tiny, darktext!60, anchor=south] at (5.3, -2.7) {Retrieval System};

\draw[dataflow, umassmaroon] (rmain) -- node[above, font=\sffamily\tiny] {$q_t$} (rret);
\draw[dataflow, darktext!50] (rret) -- node[above, font=\sffamily\tiny] {$\mathcal{D}_N$} (rrr);
\draw[dataflow, rankerpurple, rounded corners=4pt] (rrr.south) -- ++(0,-0.4) -| node[below, font=\sffamily\tiny, pos=0.18] {$o_t$ (top-$K$)} (rmain.south);

\node[font=\sffamily\scriptsize, darktext, align=center, anchor=west] (rrew) at (9.3, -1.7)
    {$r_{\text{main}},\; r_{\text{rank}}$};
\node[font=\sffamily\tiny, darktext, anchor=north] at (rrew.south) {GRPO $\to$ Main + Ranker};
\draw[dataflow, umassmaroon, dashed] (rrr) -- (rrew);

\end{tikzpicture}

\caption{Both panels share the same architecture: Main Agent $\to$ Retriever $\to$ Ranker. Top: existing approaches treat the entire retrieval system as fixed (\iceicon[0.7em]) and train only the main agent. Bottom: \methodname{} decomposes the retrieval system into a fixed dense retriever and a trainable generative ranker (\flameicon[0.7em]), jointly optimizing both the main agent and the ranker via GRPO.}
\label{fig:framework}
\vspace{-8pt}
\end{figure}
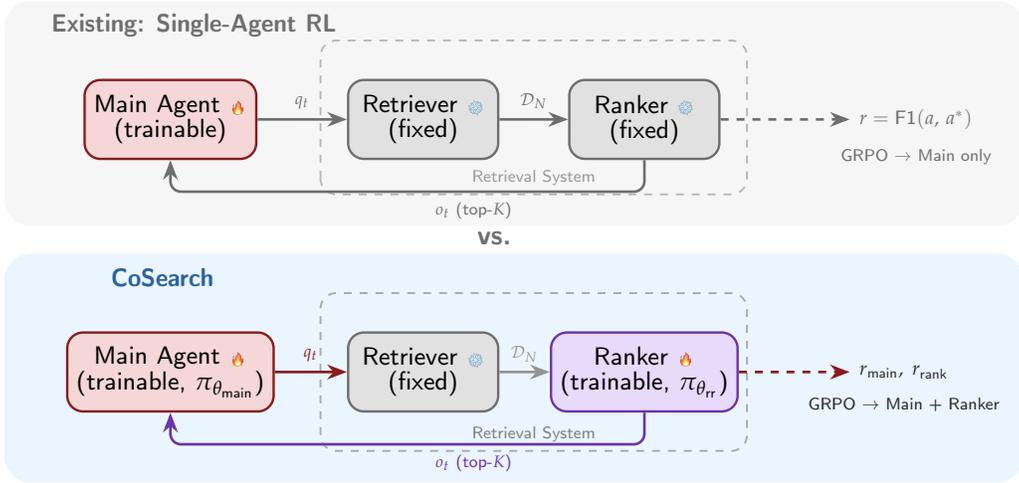

We present \methodname{}, a framework that jointly optimizes a multi-step reasoning (main) agent and a generative document ranker through reinforcement learning (RL) (see Figure~\ref{fig:framework}).
We first formalize the framework (\S\ref{sec:formulation}), describe the main agent optimization (\S\ref{sec:main-agent}), and then detail our two core technical contributions: a semantic grouping strategy for applying GRPO to the ranker (\S\ref{sec:grouping}) and a composite reward design (\S\ref{sec:reward}).

\vspace{-4pt}
\subsection{Problem Formulation}
\label{sec:formulation}

We formalize agentic search as a system comprising a reasoning (main) agent and a retrieval system. Given an initial user query $q_0$, the \emph{main agent} $\pi_{\theta_\text{main}}$ interacts with the retrieval system through a sequence of reasoning and retrieval steps to produce a trajectory:
\begin{equation}
y = \bigl(q_0,\; \tau_1, q_1, o_1,\; \tau_2, q_2, o_2,\; \dots,\; \tau_m, q_m, o_m,\; \tau_{m+1}, a\bigr),
\label{eq:trajectory}
\end{equation}
where $\tau_t$ is the reasoning thought at step $t$, $q_t$ is the sub-query generated by the main agent, $o_t$ is the observation (retrieved documents) returned by the retrieval system, and $a$ is the final answer.

In existing agentic search frameworks~\citep{search-r1,gao2025beyondTT,search-o1,zheng2025deepresearcherscalingdeepresearch}, the observation $o_t$ is provided by a fixed retrieval. In our framework, we decompose the retrieval system into two stages and make the second stage trainable: a first-stage dense retriever produces a candidate set $\mathcal{D}_N$ of $N$ documents, and a \emph{generative ranker} $\pi_{\theta_\text{gr}}$ then selects and ranks a subset $\mathcal{D}_K \subset \mathcal{D}_N$ of $K$ documents to form the observation $o_t$ (see Appendix~\ref{app:retrieval-system} for the full algorithm and concrete examples). The dense retriever remains fixed; the ranker is the trainable component. The joint optimization objective is:
\begin{equation}
\max_{\theta_\text{main},\, \theta_\text{gr}} \;\; \mathbb{E}_{T \sim \pi_{\theta_\text{main}},\, \pi_{\theta_\text{gr}}} \bigl[ r(T) \bigr],
\label{eq:joint-obj}
\end{equation}
where $r(T)$ measures the quality of the final answer. Unlike existing approaches that freeze the retrieval system and only optimize $\theta_\text{main}$, our framework makes $o_t$ a function of the trainable ranker, enabling the ranker to be optimized end-to-end by downstream answer correctness.

\vspace{-4pt}
\subsection{Main Agent Optimization}
\label{sec:main-agent}

The main agent is optimized over complete reasoning trajectories using GRPO~\citep{deepseek2025deepseekr1}. For each query $q_0$, we sample $G$ trajectories $\{y_i\}_{i=1}^G$ from the current policy. The GRPO objective is:
\begin{equation}
\mathcal{J}_\text{GRPO}(\theta_\text{main}) = \mathbb{E}_{q_0,\, \{y_i\}_{i=1}^G} \left[ \frac{1}{G} \sum_{i=1}^{G} \sum_{t=1}^{|y_i|} m_{i,t} \min\!\left(\rho_{i,t}\, \hat{A}_i,\; \text{clip}(\rho_{i,t}, 1{-}\epsilon, 1{+}\epsilon)\, \hat{A}_i\right) \right],
\label{eq:grpo-main}
\end{equation}
where $\rho_{i,t} = \pi_\theta(y_{i,t} \mid y_{i,<t}) \,/\, \pi_{\theta_\text{old}}(y_{i,t} \mid y_{i,<t})$ is the importance sampling ratio and $\hat{A}_i$ is the group-normalized advantage. Following \citet{search-r1}, we apply a loss mask $m_{i,t}$ that equals 1 for tokens generated by the main agent and 0 for retrieved document tokens.

\paragraph{Reward.}
The main agent reward evaluates both answer correctness and format compliance:
\begin{equation}
r_\text{main}(q_0, y_i) = \begin{cases}
s_\text{ans}, & \text{if } f = 1, \\[2pt]
-\alpha,      & \text{if } f = 0,
\end{cases}
\label{eq:reward-main}
\end{equation}
where $s_\text{ans}$ is the token-level F1 score between the predicted and gold answers, $f \in \{0,1\}$ indicates whether the trajectory follows the required ReAct format~\citep{yao2023reactsynergizingreasoningacting,search-r1}, and $\alpha = 0.2$ is the format penalty.

\vspace{-4pt}
\subsection{Ranker Optimization}
\label{sec:search-agent}

At each search step $t$ of the $i$-th rollout, the ranker receives a tuple $(q_0,\, q^{(i)}_t,\, \mathcal{D}_N)$, where $q_0$ is the original user query, $q^{(i)}_t$ is the sub-query issued by the main agent, and $\mathcal{D}_N$ is the top-$N$ documents retrieved by the dense retriever. The ranker first reasons about the relevance of each candidate, then selects and ranks a subset of $K$ documents to form the observation $o^{(i)}_t$. We provide the detailed prompt and input--output examples in Appendix~\ref{app:search-agent-prompt}. The ranker is trained with GRPO using the same clipped surrogate objective (Eq.~\ref{eq:grpo-main}).

However, directly applying GRPO to the ranker introduces a fundamental challenge. GRPO requires multiple responses sampled from \emph{the same input prompt} to compute group-normalized advantages. For the main agent, this grouping is natural: all $G$ trajectories originate from the same query $q_0$. For the ranker, however, the input includes the sub-query $q^{(i)}_t$, which varies across rollouts---different reasoning paths produce different sub-queries. Consequently, we cannot naively treat all ranker calls under the same $q_0$ as a single GRPO group.

A naive alternative would be to sample dedicated ranker rollouts for GRPO: for each of the $G \cdot m$ sub-queries produced by the main agent, generate $H$ independent ranker responses and continue each resulting trajectory to completion to obtain a reward. However, each ranker response alters the observation $o_t$ and consequently all downstream reasoning, producing new sub-queries that themselves require $H$ rollouts---cascading across $m$ steps into $G \cdot H^m$ total trajectories, an exponential blowup that is clearly intractable.

\paragraph{\textbf{Semantic grouping and filtering.}}
\label{sec:grouping}
We address this challenge by exploiting a key observation: although sub-queries are not identical across rollouts, many are \emph{semantically equivalent}---different reasoning paths tend to ask similar sub-questions, differing only in surface-level phrasing. For a fixed $q_0$, the $G$ rollouts produce a pool of sub-queries
$\mathcal{Q}(q_0) = \{ q_t^{(i)} \mid i \in [1,G],\; t \in [1,T_i] \}$.
We partition $\mathcal{Q}(q_0)$ into semantic groups $\mathcal{G}(q_0) = \{g_1, g_2, \dots, g_M\}$ using the following greedy procedure. We iterate through the sub-queries and assign each query $q$ to an existing group $g_m$ if its token-level F1 similarity with the group's representative query $q_m^\text{rep}$ exceeds a threshold $\delta$:
\begin{equation}
q \in g_m \iff \text{F1}_\text{token}(q,\; q_m^\text{rep}) \geq \delta,
\label{eq:grouping}
\end{equation}
where $\delta$ is a similarity threshold. Token-level F1 treats each query as a bag of tokens and computes the harmonic mean of precision and recall. If no existing group matches, a new group is created with $q$ as its representative. Groups with fewer than $k_{\min}$ members are discarded to maintain stable advantage estimation.

Each remaining group $g_m$ contains ranker calls with semantically equivalent inputs under the same $q_0$, forming a valid GRPO group. Advantages are computed within each group using standard group normalization. This strategy incurs \emph{zero additional sampling cost}: we reuse the trajectories already generated for the main agent, rather than performing separate rollouts for the ranker. We provide a detailed algorithm and an illustration in Appendix~\ref{app:grouping-alg}.

\vspace{-4pt}
\subsection{Ranker Reward Design}
\label{sec:reward}

The ranker aims to help the main agent solve a multi-step reasoning task by providing useful documents at each search step. However, reward design is non-trivial. A trajectory-level reward based on the correctness of the final answer cannot accurately assign credit to individual search steps, because each step’s contribution is mixed with subsequent retrieval decisions and with the main agent’s reasoning, making the signal noisy and unstable. A relevance-based reward also has an important limitation in our setting. Since pseudo-relevance labels are constructed by matching documents to the gold answer, they fail to supervise intermediate retrieval steps whose retrieved documents are useful for decomposition or reasoning but do not directly mention the final answer. To address these two issues, we use a composite reward that combines trajectory-level supervision with relevance-based supervision.
\paragraph{Relevance reward.}
The relevance reward $r_{\text{rel}}$ measures ranking quality using pseudo-relevance labels: a document is labeled pseudo-relevant if it contains the gold answer. We compute $r_{\text{rel}}$ as the average of Hit@$k$ over a set of cutoff values $\mathcal{K}$:
\[
r_{\text{rel}} = \frac{1}{|\mathcal{K}|} \sum_{k \in \mathcal{K}} \text{Hit@}k,
\]
where Hit@$k$ indicates whether any pseudo-relevant document appears in the top-$k$ positions of the ranker's output.
\paragraph{Main agent reward.}
The main agent reward $r_{\text{main}}$ follows the same definition as Eq.~\ref{eq:reward-main}, i.e., it is given by the token-level F1 score between the predicted and gold answers. 
Notice that Eq.~\ref{eq:reward-main} assigns a penalty $-\alpha$ when the main agent violates the required format ($f=0$). In this case, we do not propagate this penalty to the ranker. Instead, we filter out such trajectories and exclude them from reward computation and subsequent optimization.

\paragraph{Composite reward.}
Let $I_{\text{ans}} \in \{0,1\}$ indicate whether the candidate set $\mathcal{D}_N$ contains a pseudo-relevant document, and let $\gamma$ be a threshold controlling the reward composition. The final ranker reward is:
\begin{equation}
r_{\text{rank}} =
\begin{cases}
-\alpha, & f = 0, \\[4pt]
r_{\text{rel}}, & f = 1,\; I_{\text{ans}} = 1,\; r_{\text{rel}} \le \gamma, \\[2pt]
r_{\text{rel}} + r_{\text{main}}, & f = 1,\; I_{\text{ans}} = 1,\; r_{\text{rel}} > \gamma, \\[2pt]
r_{\text{main}}, & f = 1,\; I_{\text{ans}} = 0,\; r_{\text{rel}} = 0.
\end{cases}
\label{eq:reward-search}
\end{equation}

The format indicator $f$ and penalty $\alpha$ are the same as those in Eq.~\ref{eq:reward-main}. The design follows three cases. When $I_{\text{ans}} = 1$, a pseudo-relevant document exists in the candidate set. If $r_{\text{rel}}$ is low ($r_{\text{rel}} \le \gamma$), it indicates that the ranker fails to rank relevant documents effectively. In this case, we rely solely on $r_{\text{rel}}$ for supervision, since incorporating $r_{\text{main}}$ would introduce noise from the main agent's reasoning process. When $r_{\text{rel}} > \gamma$, the ranking is sufficiently strong, and we additionally incorporate $r_{\text{main}}$ to capture whether the retrieved documents contribute to the final answer. When $I_{\text{ans}} = 0$, no pseudo-relevant document exists in the candidate set, and we rely on $r_{\text{main}}$ as the only available signal. 

In sum, the composite reward $r_\text{rank}$ prioritizes ranking precision when a relevant document is available but poorly ranked, adds trajectory-level credit when the ranking is already strong, and falls back to answer correctness alone when no relevant document exists in the candidate set.

\section{Experiments}
\label{sec:experiments}
\vspace{-8pt}
\subsection{Experimental Setup}
\label{sec:setup}

\paragraph{Datasets.}
We evaluate on seven QA benchmarks spanning single-hop and multi-hop settings: PopQA~\citep{mallen2023llm_memorization}, Natural Questions (NQ)~\citep{kwiatkowski-etal-2019-natural}, TriviaQA (TQA)~\citep{joshi-etal-2017-triviaqa}, HotpotQA~\citep{hotpotqa}, 2WikiMultiHopQA (2Wiki)~\citep{2wiki}, Musique~\citep{musique}, and Bamboogle~\citep{bamboogle}. We use the test splits from Search-R1~\citep{search-r1} for consistent comparison. Performance is measured by token-level F1 score.

\paragraph{Baselines.}
We compare against Direct Inference, CoT~\citep{wei2023chainofthoughtpromptingelicitsreasoning}, RAG~\citep{rag}, Search-o1~\citep{search-o1}, Search-R1~\citep{search-r1}, ZeroSearch~\citep{sun2025zerosearch}, and two internal variants: (1)~\textbf{Retrieval Only}, which removes the ranker and directly uses the top-$K$ documents from the dense retriever; and (2)~\textbf{Fixed Ranker}, which uses a ranker fine-tuned on a relevance-based IR dataset but kept frozen during RL training.

\paragraph{Implementation details.}
For the main agent, we evaluate both Qwen2.5-7B-Instruct and Qwen2.5-3B-Instruct~\citep{qwen2} to assess performance across different model scales. The ranker in \methodname{} is based on Qwen2.5-7B-Instruct. We use the 2018 Wikipedia dump~\citep{Karpukhin2020DensePR} as the knowledge source and E5 base model~\citep{wang2024textembeddingsweaklysupervisedcontrastive} as the first-stage dense retriever. 
During rollout, each query generates $G=8$ trajectories with up to 6 search calls per trajectory. The first-stage retriever returns $N=50$ candidate passages, from which the ranker selects the top $K=5$. Training uses a learning rate of $1{\times}10^{-6}$, rollout batch size of 512, and effective batch size of 128 for training, and sampling temperature of 1.0. For semantic grouping, we set the similarity threshold to $\delta = 0.8$ and minimum group size $k_\text{min} = 3$. The composite reward threshold is $\gamma = 0.5$ and the Hit@$k$ cutoff set is $\mathcal{K} = \{1, 3, 5\}$. Further details on training data and Fixed Ranker training are provided in Appendix~\ref{app:impl}.

\vspace{-8pt}
\subsection{Main Results}
\label{sec:main-results}

\begin{table}[t]
\centering
\label{tab:main}
\vspace{4pt}
\small
\setlength{\tabcolsep}{4pt}
\begin{tabular}{l ccc cccc c}
\toprule
& \multicolumn{3}{c}{\textbf{General QA}} & \multicolumn{4}{c}{\textbf{Multi-Hop QA}} & \\
\cmidrule(lr){2-4} \cmidrule(lr){5-8}
\textbf{Method} & NQ & TriviaQA & PopQA & HotpotQA & 2Wiki & Musique & Bamboogle & \textbf{Avg.} \\
\midrule
\multicolumn{9}{l}{\textbf{Qwen2.5-7B-Instruct}} \\
Direct Inference & 0.279 & 0.504 & 0.215 & 0.292 & 0.304 & 0.118 & 0.372 & 0.298 \\
CoT              & 0.198 & 0.456 & 0.150 & 0.244 & 0.264 & 0.095 & 0.310 & 0.245 \\
RAG              & 0.420 & 0.689 & 0.387 & 0.371 & 0.244 & 0.148 & 0.296 & 0.365 \\
Search-o1        & 0.345 & 0.526 & 0.248 & 0.316 & 0.286 & 0.126 & 0.422 & 0.324 \\
Search-R1        & 0.556 & 0.681 & \textbf{0.539} & 0.576 & 0.547 & 0.272 & 0.558 & 0.533 \\
ZeroSearch       & 0.562 & 0.693 & 0.521 & 0.575 & 0.559 & 0.291 & 0.572  & 0.539 \\
\midrule
Retrieval Only   & 0.564 & 0.728 & 0.508 & 0.572 & 0.556 & 0.320 & 0.574 & 0.546 \\
Fixed Ranker   & \underline{0.578} & \underline{0.735} & 0.514 & \underline{0.578} & \underline{0.560} & \underline{0.315} & \underline{0.588} & \underline{0.553} \\
\rowcolor{umassmaroonlight}
\textbf{\methodname{}} & \textbf{0.582} & \textbf{0.747} & \underline{0.527} & \textbf{0.595} & \textbf{0.587} & \textbf{0.334} & \textbf{0.601} & \textbf{0.568} \\
\rowcolor{black!5}
\color{textgray}\textit{Oracle} & \color{textgray}0.680 & \color{textgray}0.764 & \color{textgray}0.684 & \color{textgray}0.626 & \color{textgray}0.584 & \color{textgray}0.393 & \color{textgray}0.652 & \color{textgray}0.626 \\
\midrule
\multicolumn{9}{l}{\textbf{Qwen2.5-3B-Instruct}} \\
Direct Inference &  0.198 & 0.343 & 0.168 & 0.202 & 0.249 & 0.055 & 0.068 & 0.180 \\
CoT              & 0.191 & 0.363 & 0.159 & 0.197 & 0.229 & 0.070 & 0.100 & 0.187 \\
RAG              & 0.317 & 0.478 & 0.291 & 0.371 & 0.193 & 0.110 & 0.058 & 0.223 \\
Search-o1        & 0.329 & 0.501 & 0.351 & 0.225 & 0.128 & 0.096 & 0.216 & 0.236 \\
Search-R1        & 0.458 & 0.637 & 0.460 & 0.450 & 0.397 & 0.179 & 0.414 & 0.428 \\
ZeroSearch       & 0.451 & 0.642 & 0.471 & 0.455 & 0.387 & 0.191 & 0.431 & 0.433 \\
\midrule
Retrieval Only   & 0.465 & 0.651 & 0.478 & 0.454 & 0.406 & 0.189 & 0.436 & 0.440 \\
Fixed Ranker   & \underline{0.515} & \underline{0.697} & \underline{0.483} & \underline{0.474} & \underline{0.411} & \underline{0.195} & \underline{0.448} & \underline{0.460} \\
\rowcolor{umassmaroonlight}
\textbf{\methodname{}} & \textbf{0.538} & \textbf{0.714} & \textbf{0.489} & \textbf{0.478} & \textbf{0.428} & \textbf{0.193} & \textbf{0.457} & \textbf{0.471} \\
\rowcolor{black!5}
\color{textgray}\textit{Oracle} & \color{textgray}0.548 & \color{textgray}0.749 & \color{textgray}0.612 & \color{textgray}0.589 & \color{textgray}0.510 & \color{textgray}0.308 & \color{textgray}0.589 & \color{textgray}0.558 \\
\bottomrule
\end{tabular}
\caption{Main results (token-level F1) on seven QA benchmarks. Results are grouped by main agent backbone. The \methodname{} ranker uses Qwen2.5-7B-Instruct in both settings. Best in \textbf{bold} within each scale block. Oracle (gray) promotes the answer-containing documents to rank~1, serving as an upper bound on ranking.}
\end{table}

Table~\ref{tab:main} presents the full results across all seven benchmarks. We observe that:

\paragraph{Joint training consistently outperforms all baselines.} \methodname{} achieves an average F1 of 0.568 with the 7B agent, outperforming Search-R1 by a relative 6.6\%, Retrieval Only by 4.0\% and Fixed Ranker by 2.7\%. The improvement holds across all seven benchmarks, indicating that jointly training the ranker provides a robust and general benefit. Among non-RL baselines, RAG improves over direct inference by 22.5\% (0.365 vs.\ 0.298), and Search-R1 further surpasses RAG by 46.0\% (0.533 vs.\ 0.365), confirming the value of iterative reasoning, yet \methodname{} pushes performance further by optimizing the retrieval system itself.

\paragraph{Weaker agents benefit more from retrieval improvements.} With a 3B main agent paired with a 7B ranker, \methodname{} achieves a relative 7.0\% gain over Retrieval Only and 2.4\% over Fixed Ranker. Compared to the oracle upper bound, \methodname{} closes approximately 27\% of the retrieval gap for the 7B agent and 26\% for the 3B agent, confirming that end-to-end RL optimization meaningfully narrows the performance ceiling.

\vspace{-8pt}
\subsection{Ablation Study}
\label{sec:ablation}

\begin{table}[t]
\centering
\label{tab:ablation}
\vspace{4pt}
\small
\setlength{\tabcolsep}{5pt}
\begin{tabular}{l cc c}
\toprule
\textbf{Configuration} & \textbf{Single-Hop} & \textbf{Multi-Hop} & \textbf{Avg} \\
\midrule
\textbf{\methodname{} (full)} & \textbf{0.619} & \textbf{0.529} & \textbf{0.568} \\
\midrule
\multicolumn{4}{l}{\textit{(a) Reward signal}} \\
\quad w/o $r_\text{main}$ & 0.612 & 0.508 & 0.560 \\
\quad w/o $r_\text{rel}$ & 0.591 & 0.513 & 0.552 \\
\quad Replace Hit@$k$ with nDCG@$k$ & 0.605 & 0.519 & 0.556 \\
\quad Replace Composite with LLM-as-judge & 0.608 & 0.520 & 0.558 \\
\midrule
\multicolumn{4}{l}{\textit{(b) Semantic grouping}} \\
\quad w/o semantic grouping & 0.595 & 0.503 & 0.549 \\
\midrule
\multicolumn{4}{l}{\textit{(c) Architecture}} \\
\quad Shared backbone & 0.613 & 0.518 & 0.559 \\
\quad Cross-Encoder Ranker & 0.601 & 0.501 & 0.551 \\
\quad 3B ranker & \multicolumn{3}{c}{\color{textgray}\textit{diverged (training crash)}} \\
\bottomrule
\end{tabular}
\caption{Ablation study. The main agent uses Qwen2.5-7B-Instruct. Each row changes one component of the full model. Averages are computed over single-hop and multi-hop groups separately.}
\end{table}

Table~\ref{tab:ablation} reports ablation results. On reward design, removing the trajectory-level reward $r_\text{main}$ reduces average F1 by a relative 1.4\%, while removing the ranking-quality reward $r_\text{rel}$ causes a larger drop of 2.8\%, confirming that both components provide complementary supervision: $r_\text{rel}$ supplies a direct signal on retrieval precision, while $r_\text{main}$ captures whether retrieved documents actually contribute to final answer correctness. Replacing Hit@$k$ with nDCG@$k$ reduces performance by a relative 2.1\%, and substituting the composite reward with an LLM-as-judge yields a 1.8\% drop---both confirming that our composite design is effective and computationally lightweight.

Removing the semantic grouping and filtering strategy---treating all ranker calls sharing the same initial query $q_0$ as one group regardless of sub-query similarity---results in the largest single-component degradation of 3.3\% relative (0.568 $\to$ 0.549). This confirms that mixing semantically heterogeneous sub-queries into a single GRPO group produces noisy advantage estimates that impede ranker learning, and that the filtering step described in \S\ref{sec:search-agent} is critical for stable training.

Among architectural choices, a shared backbone degrades F1 by 1.6\% due to conflicting ranking and reasoning gradients. A cross-encoder ranker drops 2.9\%: its pointwise scoring is ill-suited for direct RL optimization. Scaling the ranker to 3B causes training to diverge entirely---the smaller model cannot reliably produce valid permutation outputs over 50 candidates (see Appendix~\ref{app:search-agent-prompt}).

\vspace{-8pt}
\subsection{Analysis}
\label{sec:analysis}

\begin{figure}[t]
\centering
\includegraphics[width=0.95\textwidth]{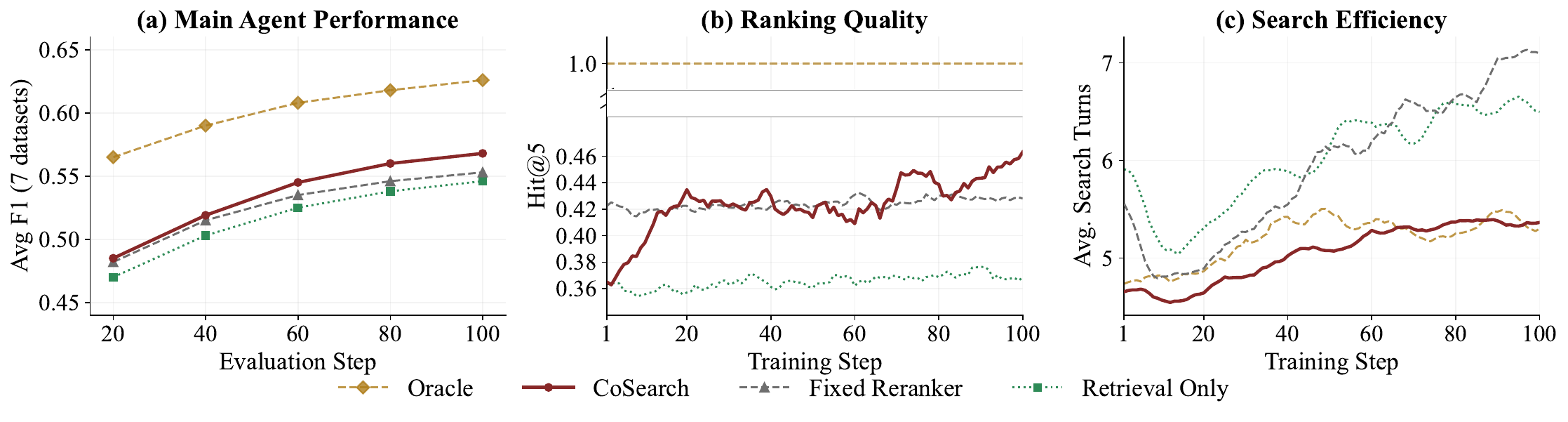}
\caption{Training and evaluation dynamics across 100 steps. (a)~Validation F1 (avg.\ over 7 datasets) every 20 steps. (b)~Ranking quality (Hit@5) at each training step; y-axis broken to show Oracle~$=1.0$ above. (c)~Average search turns per trajectory.}
\label{fig:training-dynamics}
\vspace{-12pt}
\end{figure}

Figure~\ref{fig:training-dynamics} compares all four systems across three dimensions over 100 training steps. Panels~(a) and~(b) reveal a clear positive correlation between ranking quality (Hit@5) and downstream F1, with \methodname{}'s ranker starting below the Fixed Ranker but steadily surpassing it during training. Panel~(c) shows that \methodname{} and Oracle require fewer search turns, suggesting higher-quality documents reduce the need for additional retrieval rounds.

\begin{figure}[t]
\centering
\includegraphics[width=0.95\textwidth]{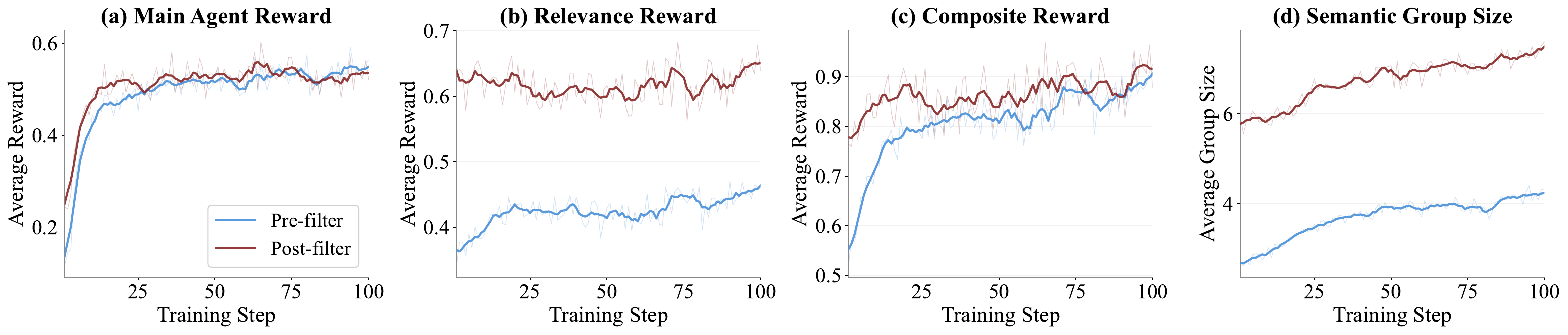}
\caption{Ranker training dynamics across 100 steps. \textbf{Pre-filter}: statistics over all semantic groups; \textbf{post-filter}: statistics over groups that pass the minimum-size filter described in \S\ref{sec:grouping}. (a)~Main agent reward. (b)~Relevance reward (Hit@5). (c)~Composite reward. (d)~Average semantic group size.}
\label{fig:ranker-training}
\vspace{-12pt}
\end{figure}

Figure~\ref{fig:ranker-training} tracks the ranker's training dynamics. In panels~(a) and~(c), pre-filter rewards eventually converge toward post-filter, suggesting the distributional gap is progressively mitigated. Panel~(b) shows pre-filter relevance reward rising steadily (from ${\sim}0.35$ to ${\sim}0.50$), indicating genuine improvement in ranking ability. Panel~(d) shows group size increasing monotonically, with post-filter roughly twice pre-filter, reflecting increasingly consistent sub-queries as the policy converges.

\begin{table}[h]
\centering
\label{tab:topk}
\vspace{4pt}
\small
\setlength{\tabcolsep}{4pt}
\begin{tabular}{c llll}
\toprule
$K$ & Retrieval Only & Fixed Ranker & \methodname{} & Oracle \\
\midrule
1  & 0.498 {\scriptsize\color{deltared}($-$8.8\%)} & 0.513 {\scriptsize\color{deltared}($-$7.2\%)} & 0.538 {\scriptsize\color{deltared}($-$5.3\%)} & 0.612 {\scriptsize\color{deltared}($-$2.2\%)} \\
3  & 0.531 {\scriptsize\color{deltared}($-$2.7\%)} & 0.539 {\scriptsize\color{deltared}($-$2.5\%)} & 0.561 {\scriptsize\color{deltared}($-$1.2\%)} & 0.631 {\scriptsize\color{deltagreen}(+0.8\%)} \\
\rowcolor{baselinegray}
\textbf{5}  & 0.546 & 0.553 & \textbf{0.568} & 0.626 \\
10 & 0.555 {\scriptsize\color{deltagreen}(+1.6\%)} & 0.559 {\scriptsize\color{deltagreen}(+1.1\%)} & 0.571 {\scriptsize\color{deltagreen}(+0.5\%)} & 0.639 {\scriptsize\color{deltagreen}(+2.1\%)} \\
\bottomrule
\end{tabular}
\caption{Effect of the number of ranked documents $K$ on Avg F1. The ranker selects the top $K$ from $N{=}50$ first-stage candidates. $K{=}5$ (bold) is the default used in Table~\ref{tab:main}. Percentages in parentheses indicate relative change from $K{=}5$ for each method.}
\end{table}

Table~\ref{tab:topk} examines how $K$ affects performance. When $K$ is reduced from 5 to 1, \methodname{} degrades by only 5.3\%, comparable to Oracle ($-$2.2\%) and far less than Retrieval Only ($-$8.8\%) and Fixed Ranker ($-$7.2\%). Notably, \methodname{} at $K{=}3$ (0.561) already surpasses Fixed Ranker at $K{=}10$ (0.559), and increasing $K$ from 5 to 10 yields diminishing returns across all methods (+0.5\% to +1.6\%).

\section*{Conclusion}
\vspace{-6pt}
In this work, we identified retrieval quality as a significant bottleneck for agentic search systems, with oracle retrieval gaps of up to +26.8\% relative F1. Motivated by this finding, we presented \methodname{}, a framework that jointly trains a reasoning agent and a generative document ranker via RL. Our semantic grouping strategy enables efficient GRPO training for the ranker without additional rollouts, and our composite reward provides both immediate ranking-quality and long-term trajectory-level signals. Experiments on seven QA benchmarks demonstrate consistent improvements, achieving +6.6\% relative F1 over Search-R1 with a 7B agent and +10.0\% with a 3B agent. Our results show that joint training of reasoning and retrieval is not only feasible but strongly performant, pointing to a key ingredient for future search agents.


\bibliography{colm2026_conference}
\bibliographystyle{colm2026_conference}

\appendix

\newpage

\section*{Appendix}
\section{Agent Prompts}
\label{app:prompts}

\subsection{Main Agent Prompt}
\label{app:main-agent-prompt}

The main reasoning agent operates in a \textsc{ReAct}-style loop with a single \texttt{search} tool.
At each turn, the agent outputs a \texttt{\small<reason>} block followed by either a
\texttt{\small<tool\_call>} block (to issue a search query) or an \texttt{\small<answer>} block
(to finalize its response).
The full prompts used across all experiments are presented below.

\begin{lstlisting}[style=promptstyle]
You are a tool-augmented research agent for wiki-based factoid question answering.

Your task is to answer questions drawn from Wikipedia-style datasets.
The final answer is evaluated using exact match (EM) or token-level F1, so it must be short and precise.

You have ONE tool available:
- search(query: string) -> returns a list of Wikipedia passages

============================================================
CRITICAL OUTPUT FORMAT (MUST FOLLOW EXACTLY)
============================================================

For EVERY assistant turn, you MUST output EXACTLY TWO TAG BLOCKS in this order:

1) <reason> ... </reason>
2) EITHER:
   (A) <tool_call> ... </tool_call>
   OR
   (B) <answer> ... </answer>

No other text is allowed outside these tags.
Do NOT output <tool_response>. The environment will provide tool results separately.

Allowed patterns:
- <reason> ... </reason>
  <tool_call> ... </tool_call>

- <reason> ... </reason>
  <answer> ... </answer>

If you violate the format, your output is invalid.

============================================================
TOOL CALL JSON SCHEMA (STRICT)
============================================================

When calling the tool, the <tool_call> block MUST contain ONLY a valid JSON object:

<tool_call>
{
  "name": "search",
  "arguments": {
    "query": "<string>"
  }
}
</tool_call>

Rules:
- "name" MUST be exactly "search"
- "arguments" MUST be an object
- "query" MUST be a single string
- Do NOT add extra keys
- Do NOT wrap JSON in Markdown
- Do NOT include comments, trailing commas, or natural language

============================================================
GENERAL TOOL USAGE
============================================================

Use the search tool whenever additional evidence would help you determine the correct answer.
If you believe you already have sufficient information to answer correctly, answer directly.

You may use multiple search calls across turns.

============================================================
SEARCH GUIDELINES
============================================================

- Write search queries that are clear and specific to what you want to confirm or find.
- After receiving evidence, reassess whether you can answer; if not, search again with a refined query.

============================================================
REASONING CONTENT REQUIREMENTS
============================================================

- Do NOT include tool JSON inside <reason>.
- Do NOT include <tool_call> or <answer> tags inside <reason>.

============================================================
ANSWER REQUIREMENTS (STRICT: SHORT ANSWER)
============================================================

Inside <answer>, you MUST:
- Output ONLY the final answer string
- Do NOT include explanations, reasoning, or extra text
- Do NOT include citations, sources, or formatting
- Use a concise canonical form (Wikipedia-style when possible)

Examples of valid answers:
- Paris
- 1997
- George Washington
- The Lord of the Rings

If the expected answer type is a person/place/organization/title/date, output only that span.
If multiple surface forms are possible, output the most standard form.

============================================================
INTEGRITY
============================================================

- Do not fabricate facts.
- If you are uncertain, use search to verify.
- If evidence is conflicting, search again with a query that resolves the conflict.

============================================================
BEGIN
============================================================

Question: {question}
\end{lstlisting}


\subsection{Ranker Prompt}
\label{app:search-agent-prompt}

The ranker receives the original user query~$q_0$, the sub-query~$q_t$ issued by the main agent
at step~$t$, and the top-$N$ candidate documents~$\mathcal{D}_N$ returned by the dense retriever.
It outputs a \texttt{\small<reason>} block containing its relevance analysis, followed by a
\texttt{\small<rerank>} block listing the selected $K$ document indices in descending order of
relevance (e.g., \texttt{\small[3] > [1] > [5] > [2]}).
The full prompt template is shown below.

\begin{lstlisting}[style=promptstyle]
You are a document ranking agent assisting a search-augmented reasoning system.

You will be given:
- An original user question
- A sub-query issued by the reasoning agent at the current search step
- A list of N candidate documents retrieved by a dense retriever,
  each labeled [1], [2], ..., [N]

Your task is to select the K most relevant documents and rank them in descending
order of relevance to help the reasoning agent answer the original question.

============================================================
CRITICAL OUTPUT FORMAT (MUST FOLLOW EXACTLY)
============================================================

You MUST output EXACTLY TWO TAG BLOCKS in this order:

1) <reason> ... </reason>
2) <rerank> ... </rerank>

No other text is allowed outside these tags.

============================================================
REASON BLOCK
============================================================

Inside <reason>, analyze the relevance of each candidate document.
Consider whether the document provides factual evidence that can help answer the
original question, taking the sub-query into account as additional context.

============================================================
RERANK BLOCK
============================================================

Inside <rerank>, list exactly K document indices in descending order of relevance:

<rerank>[id1] > [id2] > ... > [idK]</rerank>

Rules:
- Use only the provided document indices (e.g., [1], [3], [7])
- Select exactly K indices; do NOT select more or fewer
- Do NOT duplicate indices
- Do NOT include any explanation inside <rerank>

============================================================
BEGIN
============================================================

Original Question: {original_question}
Sub-Query: {sub_query}

Candidate Documents:
{documents}
\end{lstlisting}

\subsection{Ranker Input/Output Examples}
\label{app:ranker-examples}

We present two concrete examples of ranker inputs and outputs at training step~100. Each example shows the original question, the sub-query issued by the main agent at a single retrieval step, the first five of the $N{=}50$ candidate passages (truncated), and the ranker's full output comprising a relevance analysis (\texttt{<reason>}) followed by a ranked list of $K{=}5$ document indices (\texttt{<rerank>}).

\paragraph{Example 1: answer-containing document present ($I_{\text{ans}}=1$), high composite reward}
Golden answer: \textit{Harpswell}. \texttt{answer\_in\_docs}~$=\text{true}$, $r_{\text{rel}}=1.00$, $r_{\text{main}}=1.00$, $r_{\text{total}}=2.00$.

\noindent\textbf{Input (first 5 of 50 candidates shown):}
\begin{lstlisting}[style=promptstyle]
Original Question: What town has an island with a community building built in 1911-12?
Sub-Query: island with community building built 1911-12

Passages (50 total):
[1] "Bailey Island Library Hall"
Bailey Island Library Hall (locally just Library Hall) is a historic community building
at 2167 Harpswell Island Road...
[2] "Bumpkin Island"
Bumpkin Island, also known as Round Island, Bomkin Island, Bumkin Island, or Ward's Island,
is an island in the Hingham Bay area of the Bosto...
[3] "Chocorua Island Chapel"
Although the camp had fallen into disrepair by that date, the chapel continued to provide
worship services for visitors and area residents. Ev...
[4] "Gooden Grant House"
octagonal tower with turreted roof projects at the southwest corner, and a
partially-enclosed single-story porch wraps around the west and south sides...
[5] "Cherry Grove Community House and Theatre"
almost 400 other such sites have been identified as candidates. Cherry Grove Community
House and Theatre The Cherry Grove Communit...
... [45 more candidates not shown]
\end{lstlisting}

\noindent\textbf{Output:}
\begin{lstlisting}[style=promptstyle]
<reason>
The Initial Query asks for a town with an island that has a community building built in
1911-12. Passage [1] directly mentions a community building built in 1911-12 and locates
it on Harpswell Island Road, making it the top choice. Passage [48] also fits the criteria
but contains less detailed information. Passages [2], [32] discuss islands but do not
mention a community building from that period. Passage [39] describes a small island with
no matching structure.
</reason>
<rerank>[1] > [48] > [2] > [32] > [39]</rerank>
\end{lstlisting}

\paragraph{Example 2: answer-containing document absent ($I_{\text{ans}}=0$), ranker relies on $r_{\text{main}}$}
Golden answer: \textit{yes}. \texttt{answer\_in\_docs}~$=\text{false}$, $r_{\text{rel}}=0.00$, $r_{\text{main}}=1.00$, $r_{\text{total}}=1.00$.

\noindent\textbf{Input (first 5 of 50 candidates shown):}
\begin{lstlisting}[style=promptstyle]
Original Question: Are Naked Obsession and New York (1916 Film) from the same country?
Sub-Query: New York (1916 Film)

Passages (50 total):
[1] "New York (1916 film)"
New York (1916 film) New York is a lost 1916 American silent comedy drama film directed by
George Fitzmaurice and starring Florence Reed...
[2] "New York (1916 film)"
thus Oliver King becomes a benedict, and Reel 3 contains two views of a nude model.
New York is a lost 1916 American silent comedy drama film...
[3] "Lights of New York (1916 film)"
Lights of New York is a 1916 American silent drama film directed by Van Dyke Brooke.
Produced by the Vitagrap...
[4] "Lights of New York (1916 film)"
with him anyway. Lights of New York (1916 film) Lights of New York is a 1916 American
silent drama film directed by Van Dyke Brooke...
[5] "The Pride of New York"
German officer and nurse in room where bed is shown including taking nurse to room and
excluding other young woman...
... [45 more candidates not shown]
\end{lstlisting}

\noindent\textbf{Output:}
\begin{lstlisting}[style=promptstyle]
<reason>
The Initial Query asks whether Naked Obsession and New York (1916 Film) are from the same
country. The Sub-Query targets New York (1916 Film). Passages [1] and [2] directly
describe this film and confirm its American origin, making them the top choices. Passages
[3] and [4] describe a different 1916 film (Lights of New York) that is also American but
less directly relevant. Passage [5] discusses yet another unrelated film and ranks lowest.
</reason>
<rerank>[1] > [2] > [3] > [4] > [5]</rerank>
\end{lstlisting}


\section{Two-Stage Retrieval System}
\label{app:retrieval-system}

At each retrieval step $t$, the two-stage retrieval system takes the original query $q_0$, the sub-query $q_t$ issued by the main agent, and the corpus $\mathcal{C}$, and returns an observation $o_t$ consisting of $K$ ranked documents. The first stage is a fixed dense retriever that efficiently recalls $N$ candidate documents; the second stage is the generative ranker $\pi_{\theta_\text{gr}}$, which reasons about candidate relevance and outputs an explicit ranked list. Algorithm~\ref{alg:retrieval} gives the full procedure. For the prompt format and concrete input--output examples of the generative ranker, see Appendix~\ref{app:search-agent-prompt}.

\begin{algorithm}[h]
\caption{Two-Stage Retrieval at Step $t$}
\label{alg:retrieval}
\begin{algorithmic}[1]
\Require original query $q_0$, sub-query $q_t$, corpus $\mathcal{C}$, retrieval size $N$, ranking size $K$
\Ensure observation $o_t = \mathcal{D}_K$ (top-$K$ ranked documents)
\State $\mathcal{D}_N \leftarrow \textsc{DenseRetrieve}(q_t,\, \mathcal{C},\, N)$ \Comment{fixed retriever}
\State $I_{\text{ans}} \leftarrow \mathbf{1}[\text{gold answer} \in \mathcal{D}_N]$ \Comment{used for composite reward, \S\ref{sec:reward}}
\State $\textit{prompt} \leftarrow \textsc{BuildPrompt}(q_0,\, q_t,\, \mathcal{D}_N,\, K)$
\State $\textit{output} \leftarrow \pi_{\theta_\text{gr}}(\textit{prompt})$ \Comment{generative ranker produces ranked list}
\State $\mathcal{D}_K \leftarrow \textsc{ParseRanking}(\textit{output},\, \mathcal{D}_N,\, K)$ \Comment{extract top-$K$ documents}
\State \Return $\mathcal{D}_K$
\end{algorithmic}
\end{algorithm}


\section{Semantic Grouping Algorithm}
\label{app:grouping-alg}

\begin{figure}[h]
\centering
\includegraphics[width=\linewidth]{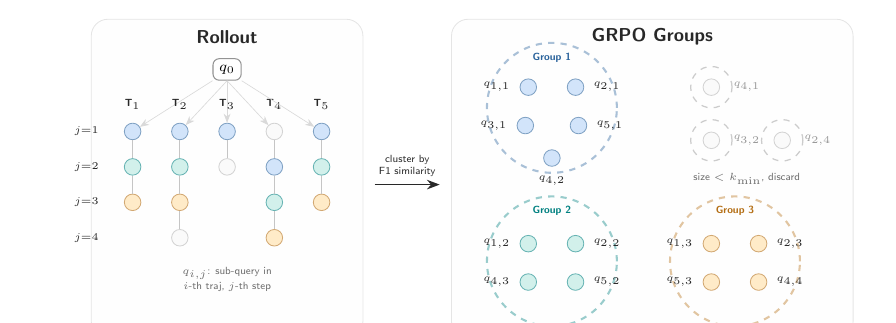}
\caption{An illustration of semantic grouping for GRPO training, shown for $G{=}5$ rollout trajectories with up to 5 search steps each. Left: each dot is a sub-query $q_{i,j}$ (trajectory $i$, step $j$) colored by group membership. Right: all sub-queries are clustered by token-level F1 similarity into GRPO groups; gray singletons (size $< k_{\min}$) are discarded before optimization.}
\label{fig:grouping-alg}
\end{figure}

Applying GRPO to the ranker requires grouping ranker calls with the same input prompt. Since sub-queries vary across rollouts, we perform a two-level grouping: first split by whether the candidate set contains the gold answer ($I_{\text{ans}}$), then cluster within each split by token-level F1 similarity. Figure~\ref{fig:grouping-alg} illustrates the procedure for a simplified example with $G{=}5$ rollouts and up to 5 search steps per trajectory; in our experiments we use $G{=}8$ rollouts with up to 6 search steps. Algorithm~\ref{alg:grouping} gives the full procedure.

\begin{algorithm}[h]
\caption{Semantic Grouping for Ranker GRPO Training}
\label{alg:grouping}
\begin{algorithmic}[1]
\Require ranker calls $\mathcal{O}$ (from $G$ rollouts of $q_0$), similarity threshold $\delta$, minimum group size $k_{\min}$
\Ensure UID-labeled ranker calls ready for GRPO
\State \textbf{// Step 1: Split by answer availability}
\State Partition $\mathcal{O}$ into $\mathcal{O}_{\text{easy}}$ ($I_{\text{ans}}=1$) and $\mathcal{O}_{\text{hard}}$ ($I_{\text{ans}}=0$)
\For{each split $\mathcal{B} \in \{\mathcal{O}_{\text{easy}},\, \mathcal{O}_{\text{hard}}\}$}
    \State \textbf{// Step 2: Greedy cluster by token-level F1}
    \State $\textit{clusters} \leftarrow \{\}$ \Comment{cluster\_id $\to$ representative sub-query}
    \For{each call $o \in \mathcal{B}$}
        \State $q \leftarrow \textsc{Normalize}(o.\text{sub\_query})$
        \State $\textit{matched} \leftarrow \text{False}$
        \For{each $(c,\, q^{\text{rep}}) \in \textit{clusters}$}
            \If{$\text{F1}_{\text{token}}(q,\, q^{\text{rep}}) \geq \delta$}
                \State assign $o \to$ cluster $c$;\quad $\textit{matched} \leftarrow \text{True}$;\quad \textbf{break}
            \EndIf
        \EndFor
        \If{not $\textit{matched}$}
            \State create new cluster with $q$ as representative;\quad assign $o \to$ new cluster
        \EndIf
    \EndFor
    \State \textbf{// Step 3: Discard small groups}
    \State Remove all clusters with $|\text{cluster}| < k_{\min}$
    \State \textbf{// Step 4: Assign UIDs}
    \For{each surviving cluster $c$}
        \State $\text{uid} \leftarrow \texttt{\{main\_uid\}\_\{easy|hard\}\_cluster\_\{}c\texttt{\}}$
        \State assign uid to all $o \in c$
    \EndFor
\EndFor
\end{algorithmic}
\end{algorithm}


\section{Additional Implementation Details}
\label{app:impl}

\paragraph{Training data.}
The RL training set consists of 51,200 questions drawn from four datasets: Natural Questions (20,480 questions, 40\%), HotpotQA (14,220, 28\%), Musique (9,000, 18\%), and 2WikiMultiHopQA (7,500, 15\%). All questions are sampled from each dataset's official training split. For evaluation, we use the official test split of each benchmark, or the evaluation split when no test split is available.

\paragraph{Fixed Ranker training.}
The Fixed Ranker baseline is trained as follows. We randomly sample 50K queries from MS~MARCO and 50K from Natural Questions, then use \texttt{e5-base} to retrieve the top-50 documents for each query. Queries for which no relevant document appears in the top-50 are discarded, yielding 92,160 training queries. The ranker is trained with RL using the same Hit@$\{1,3,5\}$ composite reward and the same generative listwise output format as the jointly trained ranker in \methodname{}.


\section{Semantic Grouping Examples}
\label{app:grouping}

Tables~\ref{tab:grouping-examples-p1} and~\ref{tab:grouping-examples-p2} present 20 sampled initial queries from the step-100 rollout along with their sub-queries after semantic grouping and filtering. For each initial query, the $G{=}8$ rollout trajectories produce a pool of sub-queries; these are clustered by token-level F1 similarity with threshold $\delta{=}0.8$. Column~\textbf{C} indicates the cluster index assigned to each sub-query. The table illustrates that multi-hop questions naturally produce multiple distinct semantic clusters (e.g., one cluster for each sub-question in a two-hop chain), while simpler questions tend to produce a single large cluster of near-paraphrases.


\begin{table}[p]
\centering
\small
\setlength{\tabcolsep}{5pt}
\renewcommand{\arraystretch}{1.2}
\label{tab:grouping-examples-p1}
\begin{tabular}{p{4.2cm} c p{7.5cm}}
\toprule
\textbf{Initial Query} & \textbf{C} & \textbf{Sub-query} \\
\midrule
\multirow{4}{4.2cm}{who has played the most state of origins} & 0 & who has played the most state of origins \\
 & 1 & current record holder for most State of Origin appearances \\
 & 2 & most State of Origin appearances \\
 & 2 & most appearances in State of Origin \\
\midrule
\multirow{3}{4.2cm}{what is the host nation of the 2004 summer olympic g...} & 0 & host nation of 2004 Summer Olympic Games \\
 & 1 & host nation of 2004 Summer Olympics \\
 & 1 & host nation of the 2004 Summer Olympics \\
\midrule
\multirow{5}{4.2cm}{What is the birth city of Blessed John the Fool-For-...} & 0 & capital of former Soviet Union and country with AGT show \\
 & 0 & capital of the former Soviet Union and country with AGT \\
 & 1 & Blessed John the Fool-For-Christ birth city \\
 & 1 & Blessed John the Fool-For-Christ birth city Azerbaijan \\
 & 1 & Blessed John the Fool-For-Christ birth city Baku Azerbaijan \\
\midrule
\multirow{3}{4.2cm}{What race is the majority of the population of the c...} & 0 & The Unbeatables I production country \\
 & 1 & The Unbeatables I \\
 & 1 & The Unbeatables I produced \\
\midrule
\multirow{4}{4.2cm}{Who plays the writer who mentioned The Angel Pub in ...} & 0 & Charles Dickens The Angel Pub \\
 & 0 & Charles Dickens The Angel Pub writings \\
 & 0 & Charles Dickens mentioned The Angel Pub \\
 & 1 & The Man Who Invented Christmas \\
\midrule
\multirow{2}{4.2cm}{Are Naked Obsession and New York (1916 Film) from th...} & 0 & Naked Obsession \\
 & 1 & New York (1916 Film) \\
\midrule
\multirow{2}{4.2cm}{Standard Chartered Bank, who sponsors the annual Hon...} & 0 & Standard Chartered Bank headquarters \\
 & 1 & Standard Chartered Bank sponsors Hong Kong Marathon \\
\midrule
\multirow{6}{4.2cm}{on what day of the year is the sun near the star reg...} & 0 & when is the sun near Regulus (Alpha Leo) on the calendar \\
 & 0 & when is the sun near Regulus (Alpha Leo) on the exact date \\
 & 0 & when is the sun near Regulus (Alpha Leo) on the year \\
 & 1 & day of the year when the sun is near Regulus (Alpha Leo) \\
 & 1 & day of the year when the sun is near Regulus (alpha leo) \\
 & 1 & exact day of the year when the Sun is near Regulus (Alpha Leo) \\
\midrule
\multirow{4}{4.2cm}{who did the congress send to london as a minister in...} & 0 & who did the congress send to london as a minister in 1784 \\
 & 1 & who was sent as a minister to london by the congress in 1784 \\
 & 1 & who was sent as a minister to london in 1784 \\
 & 1 & who was sent as minister to london by congress in 1784 \\
\midrule
\multirow{5}{4.2cm}{In what season of the 2017 year is the Netflix Germa...} & 0 & Dark (TV series) 2017 release season \\
 & 0 & Dark (TV series) release schedule 2017 \\
 & 1 & Dark (Netflix German series) season release \\
 & 1 & Dark Netflix German series 2017 release season \\
 & 1 & Dark Netflix German series release season \\
\bottomrule
\end{tabular}
\caption{Sub-queries after semantic grouping and filtering (Part~1 of~2, training step~100). For each initial query, sub-queries from $G{=}8$ rollouts are clustered by token-level F1 similarity ($\delta{=}0.8$). Column \textbf{C} denotes the cluster index.}
\end{table}

\begin{table}[p]
\centering
\small
\setlength{\tabcolsep}{5pt}
\renewcommand{\arraystretch}{1.2}
\label{tab:grouping-examples-p2}
\begin{tabular}{p{4.2cm} c p{7.5cm}}
\toprule
\textbf{Initial Query} & \textbf{C} & \textbf{Sub-query} \\
\midrule
\multirow{4}{4.2cm}{What part did The King of Hollywood play in China Seas?} & 0 & The King of Hollywood China Seas \\
 & 0 & The King of Hollywood China Seas part \\
 & 1 & Clark Gable role in China Seas \\
 & 1 & Clark Gable's role in China Seas \\
\midrule
\multirow{3}{4.2cm}{What is the Izzo (H.O.V.A.) performer's record label?} & 0 & Izzo (H.O.V.A.) performer record label \\
 & 1 & Jay-Z current record label \\
 & 1 & Jay-Z record label \\
\midrule
\multirow{6}{4.2cm}{Where was the place of death of the director of film...} & 0 & director of film When A Man Sees Red 1934 \\
 & 0 & director of film When A Man Sees Red 1934 Pursued \\
 & 0 & director of film When A Man Sees Red 1934 version \\
 & 1 & Frank Ellis death place \\
 & 1 & Frank Ellis place of death \\
 & 1 & place of death Frank Ellis \\
\midrule
\multirow{2}{4.2cm}{Who was born later, Dani Pacheco or Agnė Čepelytė?} & 0 & Dani Pacheco birth year \\
 & 1 & Agnė Čepelytė birth year \\
\midrule
\multirow{3}{4.2cm}{Who is the operator of Embassy of Northern Cyprus in...} & 0 & Embassy of Northern Cyprus in Istanbul operator \\
 & 0 & Embassy of Northern Cyprus in Kemerhisar operator \\
 & 0 & Embassy of Northern Cyprus in Samsun operator  \\
\midrule
\multirow{3}{4.2cm}{What town has an island with a community building bu...} & 0 & island with community building built 1911-12 \\
 & 0 & town with an island and a community building built in 1911-12 \\
 & 0 & town with island and community building built in 1911-12 \\
\midrule
\multirow{2}{4.2cm}{The M66 is a motorway in Lancashire and Greater Manc...} & 0 & M66 motorway in Lancashire and Greater Manchester \\
 & 0 & M66 motorway in Lancashire and Greater Manchester, England \\
\midrule
\multirow{3}{4.2cm}{who plays the queen of hearts in alice and wonderland} & 0 & who plays the Queen of Hearts in Alice and Wonderland \\
 & 0 & who plays the Queen of Hearts in Alice in Wonderland \\
 & 0 & who plays the Queen of Hearts in recent Alice in Wonderland films \\
\midrule
\multirow{1}{4.2cm}{What nationality is the director of film Porky'S Rev...} & 1 & James Komack nationality \\\\
\midrule
\multirow{3}{4.2cm}{What presenter of Market Kitchen has won a Guild of ...} & 0 & Market Kitchen presenter Guild of Food Writers award \\
 & 0 & Market Kitchen presenter who won Guild of Food Writers award \\
 & 0 & Market Kitchen presenter won Guild of Food Writers award \\
\bottomrule
\end{tabular}
\caption{Sub-queries after semantic grouping and filtering (Part~2 of~2, training step~100). For each initial query, sub-queries from $G{=}8$ rollouts are clustered by token-level F1 similarity ($\delta{=}0.8$). Column \textbf{C} denotes the cluster index.}
\end{table}


\section{Oracle Retrieval Construction}
\label{app:oracle}

The oracle retrieval experiment (Table~\ref{tab:oracle}) measures the performance gap caused by imperfect retrieval. Unlike post-hoc analysis, the oracle trajectory is constructed \emph{online} during rollout: the agent's reasoning at each step is conditioned on the (potentially modified) observation, so improved retrieval at step $t$ influences all subsequent reasoning and sub-queries.

Concretely, given an initial query $q_0$ with gold answer $a^*$, the agent generates a trajectory step by step following Eq.~\ref{eq:trajectory}. At each step $t$, the agent produces a reasoning thought $\tau_t$ and sub-query $q_t$, and the dense retriever returns a candidate set $\mathcal{D}_N$. Let $\mathcal{D}^+ = \{d \in \mathcal{D}_N \mid a^* \subseteq d\}$ denote the documents in $\mathcal{D}_N$ that contain the gold answer. The observation presented to the agent is:
\begin{equation}
o_t =
\begin{cases}
\mathcal{D}^+ \,\|\, \text{top-}(K - |\mathcal{D}^+|) \text{ from } \mathcal{D}_N \setminus \mathcal{D}^+, & \text{if } \mathcal{D}^+ \neq \emptyset, \\[4pt]
\text{top-}K \text{ from } \mathcal{D}_N, & \text{if } \mathcal{D}^+ = \emptyset,
\end{cases}
\end{equation}
where $\|$ denotes concatenation. When documents matching the gold answer exist in the candidate set, we promote them to the top positions and fill the remaining slots with the highest-ranked non-matching documents. For intermediate sub-queries where no retrieved document matches the gold answer, we use the default retriever ranking unchanged. The agent then observes $o_t$, continues reasoning, and repeats until it produces a final answer $a$. Algorithm~\ref{alg:oracle} provides the full procedure.

\begin{algorithm}[h]
\caption{Oracle Retrieval Rollout}
\label{alg:oracle}
\begin{algorithmic}[1]
\Require Initial query $q_0$; gold answer $a^*$; main agent $\pi_{\theta_\text{main}}$; candidate set size $N$; output size $K$
\Ensure Oracle trajectory $y^{\text{oracle}}$
\State $t \gets 0$
\Repeat
    \State $t \gets t + 1$
    \State $\tau_t, q_t \gets \pi_{\theta_\text{main}}(q_0, \tau_1, q_1, o_1, \dots, \tau_{t-1}, q_{t-1}, o_{t-1})$ \Comment{agent reasons and generates sub-query}
    \State $\mathcal{D}_N \gets \text{DenseRetrieve}(q_t, N)$
    \State $\mathcal{D}^+ \gets \{d \in \mathcal{D}_N \mid a^* \subseteq d\}$ \Comment{gold-answer-matching documents}
    \If{$\mathcal{D}^+ \neq \emptyset$}
        \State $o_t \gets \mathcal{D}^+ \,\|\, \text{top-}(K - |\mathcal{D}^+|) \text{ from } \mathcal{D}_N \setminus \mathcal{D}^+$ \Comment{promote to top}
    \Else
        \State $o_t \gets \text{top-}K \text{ from } \mathcal{D}_N$ \Comment{keep default ranking}
    \EndIf
\Until{agent outputs final answer $a$}
\State \Return $y^{\text{oracle}} = (q_0, \tau_1, q_1, o_1, \dots, \tau_m, q_m, o_m, \tau_{m+1}, a)$
\end{algorithmic}
\end{algorithm}

\section{Search Turn Distribution}
\label{app:search-turns}

Table~\ref{tab:search-turns} shows the full search turn distribution. \methodname{} and Oracle concentrate 91\% and 87\% of questions at 1--2 turns respectively, while Fixed Ranker pushes 55\% of questions to 3 turns, requiring significantly more retrieval rounds on average (2.62 vs.\ 1.67). This confirms that higher-quality ranking reduces the number of retrieval steps needed to resolve a question.

\begin{table}[t]
\centering
\small
\setlength{\tabcolsep}{5pt}
\label{tab:search-turns}
\begin{tabular}{l ccccccc c}
\toprule
\textbf{Model} & \textbf{0} & \textbf{1} & \textbf{2} & \textbf{3} & \textbf{4} & \textbf{5} & \textbf{6} & \textbf{Avg} \\
\midrule
\textit{Oracle} & 0.0 & 49.3 & 38.2 & 9.8 & 2.3 & 0.3 & 0.1 & 1.66 \\
\rowcolor{umassmaroonlight}\textbf{CoSearch} & \textbf{0.0} & \textbf{42.3} & \textbf{48.9} & \textbf{8.1} & \textbf{0.7} & \textbf{0.0} & \textbf{0.0} & \textbf{1.67} \\
Fixed Reranker & 0.0 & 4.2 & 35.7 & 54.5 & 5.6 & 0.0 & 0.0 & 2.62 \\
Retrieval Only & 0.0 & 33.2 & 40.4 & 16.3 & 7.9 & 1.8 & 0.3 & 2.05 \\
\bottomrule
\end{tabular}
\caption{Search turn distribution (\% of questions) at validation step~100. \textbf{Avg} is the mean number of search turns per question.}
\end{table}

\end{document}